\documentclass{article}

\usepackage{PRIMEarxiv}

\usepackage[utf8]{inputenc} 
\usepackage[T1]{fontenc}    
\usepackage{hyperref}       
\usepackage{subfigure}
\usepackage{url}            
\usepackage{booktabs}       
\usepackage{amsfonts}       
\usepackage{nicefrac}       
\usepackage{microtype}      
\usepackage{lipsum}
\usepackage{fancyhdr}       
\usepackage{graphicx}       
\usepackage{authblk}
\graphicspath{{media/}}     
\usepackage{makecell}
\title{Scalable Object Detection on Embedded Devices using Weight Pruning and Singular Value Decomposition
}
\newcommand\correspondingauthor{\thanks{Corresponding author.}}
\author{\textbf{Dohyun Ham}$^\dagger$}
\author{\textbf{Jaeyeop Jeong}$^\dagger$}
\author{\textbf{June-Kyoo Park}$^\dagger$}
\author{\textbf{Raehyeon Jeong}$^\dagger$}
\author{\textbf{Seungmin Jeon}$^\dagger$}
\author{\textbf{Hyeongjun Jeon}}
\author{\textbf{Yewon Lim}$^\dagger$\correspondingauthor}
  
\affil{
  ROBOIN \vspace{8pt}\\
  Yonsei University \\
  Seoul, Republic of Korea\\
  \texttt{\{mango3354, jeongjaeyeop, jkp, raehy19, \\algorhythm, hyeongjun, ga06033\}@yonsei.ac.kr}
}

\begin{document}
\maketitle

\def\thefootnote{$\dagger$}\footnotetext{These authors contributed equally to this work}

\begin{figure}[h!]
    \centering
    \includegraphics[width=0.45\textwidth]{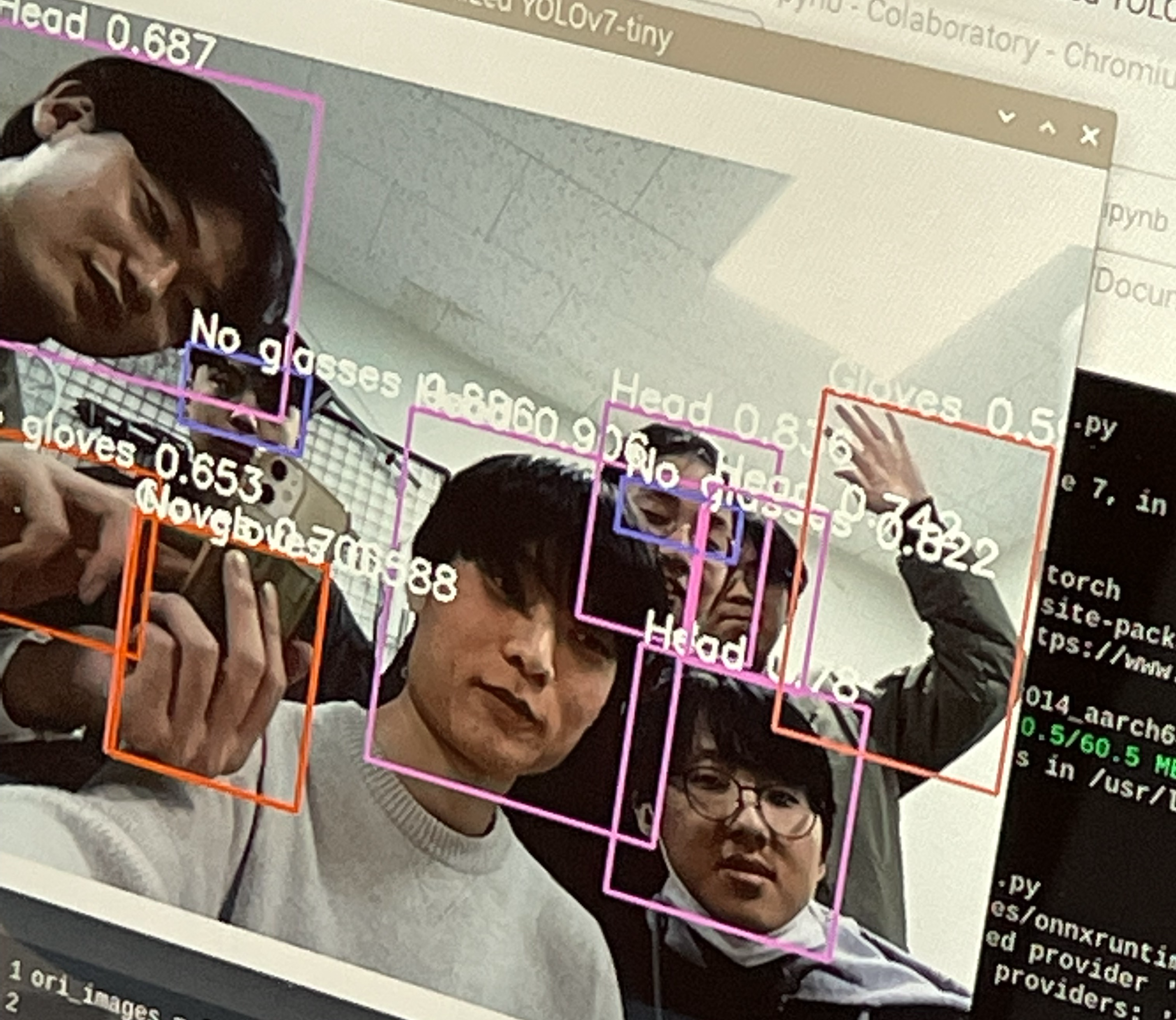}
    \caption{Real-time object detection inference on Raspberry Pi 4 using YOLOv7 model and camera module.}
    \label{fig:result}
\end{figure}

\begin{abstract}
This paper presents a method for optimizing object detection models by combining weight pruning
 and singular value decomposition (SVD). The proposed method was evaluated on a custom dataset of 
 street work images obtained from \url{https://universe.roboflow.com/roboflow-100/street-work}. 
 The dataset consists of 611 training images, 175 validation images, and 87 test images with 7 classes. 
 We compared the performance of the optimized models with the original unoptimized model in terms of frame rate, 
 mean average precision (mAP@50), and weight size. The results show that the weight pruning + SVD model 
 achieved a 0.724 mAP@50 with a frame rate of 1.48 FPS and a weight size of 12.1 MB, outperforming 
 the original model (0.717 mAP@50, 1.50 FPS, and 12.3 MB). Precision-recall curves were also plotted 
 for all models. Our work demonstrates that the proposed method can effectively optimize object 
 detection models while balancing accuracy, speed, and model size.
\end{abstract}


\section{Introduction}
Due to the limitations in battery capacity, physical volume, privacy, and latency, it is infeasible to deploy powerful machine learning models, such as YOLOv3 with more than 65 million parameters, in mobile applications such as autonomous vehicles or thermal imaging systems. Object detection is a fundamental technique in computer vision that enables a software system to detect and locate objects from an image or video stream. The key characteristic of object detection is its ability to identify the class of the object and its location in the image or video stream. The location is typically denoted by a bounding box around the object, while the object's class can be inferred from pre-trained weights associated with the bounding box.

Real-time object detection aims to perform object detection in real-time by predicting and locating objects on-the-fly. It has numerous applications including autonomous vehicles, face recognition, letter detection, and risk detection. For instance, Google Lens, first released in 2017, uses object detection to identify objects in an image or video stream. The Snapchat app can detect whether an object is an animal or a person, and applies relevant filters accordingly. Additionally, self-driving Tesla vehicles and Apple iPhone's Photo Cutout feature both rely on real-time object detection to operate.

Existing models often have a high number of parameters, making them difficult to deploy on resource-constrained hardware like the Raspberry Pi. Attempts to run these models on Raspberry Pi have resulted in low frame rates and truncated screens. To mitigate these issues, techniques like weight pruning and SVD have been proposed.

Weight pruning is a lightweight technique that selectively removes insignificant parameters from the model's weights to increase inference speed and generalization performance. However, this approach can lead to a loss of information and reduced hardware acceleration efficiency. Singular Value Decomposition (SVD) is another technique that decomposes a matrix into a specific structure and diagonalizes it. Unlike eigenvalue decomposition, SVD can be applied to all $m\times n$ matrices, irrespective of their shape.

This paper aims to implement and evaluate weight pruning and SVD as optimization techniques for a popular object detection model, YOLO-v7.
\section{Related Work}
\label{sec:relatedwork}
To infer computer vision models in real-time from resource-constrained devices, one approach is to
 design an efficient neural network structure that reduces memory and computation requirements. 
 GoogleNet\cite{GoogleNet} and SqueezeNet\cite{SqueezeNet} achieved this by reducing the number of
  parameters through the use of 1x1 convolution kernels instead of the standard 3x3 convolution kernel. 
  SqueezeNet was able to reduce the model size by 50 times compared to AlexNet while exceeding the accuracy
   of AlexNet. In MobileNet\cite{MobileNet}, the convolutional process was further reduced by decomposing 
   it into Depthwise convolution and Pointwise convolution. These attempts at efficient small model design
    are crucial, but they do have limitations. To overcome these limitations, additional methods such as model 
    quantization, pruning, and SVD have been devised for better compression in terms of efficiency.

\subsection{Quantization}
Model quantization is an effective compression technique that reduces the precision of weights and activations to reduce the model size. This technique involves converting high-precision data, such as 32-bit floating-point numbers, into low-precision data, such as 8-bit integers. While post-training quantization is the most common approach, it may lead to accuracy degradation in small models, particularly when there is a large deviation in the weight range of each channel or when an outlier weight exists.

To address these issues, quantization-aware training (QAT) has been developed, which simulates the quantization 
effect during the net propagation process of training \cite{QAT}. During QAT, all weights and biases are stored 
as floating-point numbers, and backpropagation is performed as usual. At the time of inference, the weights and 
biases are converted into 8-bit integers and used in calculations.

In addition to 8-bit integer quantization, it is also possible to use even lower bit widths to represent weights 
and activations. For example, Ternary Weight Networks (TWN) use -1, 0, and 1 bits to represent weights \cite{TWN},
 while Binary Neural Networks (BNN) use -1 and 1 bits \cite{BNN}. XNOR-Net is a model that uses binary weight and 
 input representations, which simplifies the convolution process to a scaling operation after the XNOR Gate operation 
 \cite{XNOR}. This results in a 58x speedup compared to conventional convolution while saving 32x more memory. While 
 the efficiency of these bit operators can be applied to specific custom hardware, it may not be applicable to all real hardware.

In summary, bit quantization techniques offer significant benefits in terms of model size and computational efficiency. However, their applicability to real hardware may be limited. Thus, researchers must carefully consider the trade-offs between model size, computational efficiency, and hardware constraints when choosing a quantization technique for their specific use case.

\subsection{Pruning}

Model pruning is a widely used technique in deep learning that reduces the size and computational complexity of neural networks by removing redundant or less important parameters. Two common methods for pruning are weight pruning, which removes small magnitude weights, and structured pruning, which removes entire filters or channels. While weight pruning requires additional software or hardware to support the resulting sparse matrix, structured pruning can be easily inferred and is more practical for deployment without specific processing.

Structured pruning involves marking certain structures of the neural network for removal based on their importance scores, 
which are determined by specific criteria. One such criterion is the L1-norm, which measures the importance of each filter,
 channel, or layer by the sum of the absolute values of its weights \cite{Pruning}. For example, when filters were pruned 
 in the order in which the L1-norm values of each filter were small, 64\% of the weights were removed from the VGG-16 model, 
 with no significant difference in accuracy between the unpruned and pruned models. Various methods for conducting model pruning 
 are being studied, including those that combine pruning with other techniques such as quantization and knowledge distillation. 

\subsection{YOLO-v7 }

YOLO, short for "You Only Look Once", is a widely used family of algorithms for real-time object detection. Object detection models are typically classified into two categories: one-stage detectors and two-stage detectors. One-stage detectors are designed to perform object localization and classification simultaneously, while two-stage detectors perform these tasks in two separate steps. One popular one-stage detector is YOLO, which divides the input image into a grid of cells and performs multi-class classification and bounding box regression for each cell. This approach allows YOLO to efficiently detect objects in real-time, making it well-suited for applications that require fast and accurate detection. On the other hand, two-stage detectors typically involve a region proposal network (RPN) that generates candidate object regions, followed by a separate classification and localization step. Although two-stage detectors may be slower than one-stage detectors, they often achieve higher accuracy and are commonly used in applications that require high precision object detection. Overall, the choice of object detection model depends on the specific requirements of the application, such as the desired trade-off between speed and accuracy.

YOLOv7 \cite{yolov7}, the latest iteration of this architecture, is capable of real-time object detection. The YOLOv7 
architecture relies on a convolutional neural network to extract features from input images and predict the bounding 
boxes and class probabilities for each object in the image.

The YOLOv7 architecture is composed of three main components: a backbone, a neck, and a head. 
The backbone of the YOLOv7 architecture contains a computational block called Extended Efficient Layer Aggregation Network
(E-ELAN), which builds on the existing ELAN by using expand, shuffle, and merge cardinality to enhance the learning ability 
of the network without disrupting the original gradient path. Group convolution is used to expand the channel and cardinality 
of computational blocks, and feature maps are shuffled into groups before being merged together. This approach allows 
for continuous improvement in the network's ability to detect objects in real-time. The neck of the architecture is 
responsible for gathering the feature maps that are extracted by the Backbone and using them to create feature pyramids. 
The head is responsible for producing the final model outputs, and YOLOv7 implements a technique called Deep Supervision
\cite{Supervision} to allow for multiple heads to be used during training. The lead head is responsible for the final output, 
while an auxiliary head is used to assist with training in the middle layers of the network.

Among the various iterations of YOLOv7, the YOLOv7-tiny is an optimized model designed for edge devices. In comparison to other versions, the edge-optimized YOLOv7-tiny model utilizes the Leaky ReLU activation function, while other models implement the SiLU activation function. The Leaky ReLU activation function allows for a small, non-zero output when the input is negative, which can prevent dead neurons and improve the performance of the model. In contrast, the SiLU activation function applies a sigmoid function to the input, which smooths out the outputs and can provide better gradient propagation. However, the edge-optimized YOLOv7-tiny model has been specifically designed for edge computing, where hardware limitations may require more efficient and compact models. By utilizing the Leaky ReLU activation function, the model can maintain a high level of accuracy while minimizing computational resources.

\subsection{SVD}

SVD (Singular Value Decomposition) is a mathematical technique used in linear algebra to decompose a matrix into three smaller matrices: a left singular vector matrix, a singular value diagonal matrix, and a right singular vector matrix. Truncated SVD is a variant of SVD that involves retaining only a subset of the singular values and their corresponding singular vectors. This is achieved by setting a threshold below which the singular values and their corresponding singular vectors are discarded.

In neural network compression, Truncated SVD can be used to reduce the size of weight matrices and compress the network. 
By keeping only the top k singular values and their corresponding singular vectors, a smaller compressed weight matrix can
 be obtained that can replace the original weight matrix. When SVD was performed on the model in \cite{SVDModel}, the 
 speed of the bottleneck convolution operation in the first layer increased by three times, and the number of weight 
 parameters in the fully connected layer was reduced by five to 13 times compared to the previous method with no 
 significant decrease in accuracy \cite{SVD}.

SVD is independent of other methods used for efficient evaluation, such as quantization or pruning. Therefore, it has the potential to be used in combination with other methods to achieve additional benefits.

\section{Methodology}

In this work, we used the YOLO-v7 object detection architecture available on \url{https://github.com/WongKinYiu/yolov7} and deployed it on a Raspberry Pi 4 Model B (4GB RAM). YOLO-v7 is a state-of-the-art object detection model that is highly efficient in terms of both speed and accuracy. We chose to use the Raspberry Pi 4 due to its relatively low cost and compact size, making it an ideal platform for implementing object detection models in resource-constrained environments.

\noindent{\bf Dataset}\\
In this work, we used a publicly available dataset for object detection\cite{street-work_dataset}, which consists of 
611 training images, 175 validation images, and 87 test images. The dataset includes 7 classes of objects, including 
cars, pedestrians, and bicycles.

To preprocess the images, we resized them to a fixed size of 640x640 pixels and normalized the pixel values to be
 between 0 and 1. We also randomly applied data augmentation techniques during training, such as random cropping,
  translations, and adjusting brightness and contrast.

The dataset was split into three parts: training, validation, and testing, with a ratio of approximately 70:20:10. 
We used the training set to train our models, the validation set to tune hyperparameters and prevent overfitting, 
and the test set to evaluate the performance of our models.

\noindent{\bf Pruning}\\
In this study, we applied L1 pruning to our object detection model to reduce its size and improve its efficiency for
 deployment on resource-constrained devices. L1 pruning is a popular technique for compressing deep neural networks by
  removing the least important weights in the network. This is achieved by identifying the 30\% of the total elements 
  of weight tensors with the lowest L1 norms and setting them to zero.

The motivation for using L1 pruning is based on the assumption that a small fraction of the weights in a neural network 
are responsible for the majority of its representational power. By removing the least important weights, we can reduce 
the model size and computational requirements without significantly affecting its performance.

\noindent{\bf Low-Rank SVD}\\
We employed singular value decomposition (SVD) as a model optimization technique to reduce the size of the object detection 
model. We first reshaped the weight tensor of each convolutional layer in the model into a two-dimensional matrix. 
We reshaped the matrix in a way that makes the number of rows of the matrix be the largest integer that is a divisor 
of the number of the total elements of the weight tensor. We intended to produce a square matrix as much as possible 
to make the SVD process more efficient.

We then applied a low-rank SVD to this matrix, which factorizes the matrix into three separate matrices: a left singular 
matrix, a diagonal singular value matrix, and a right singular matrix. We truncated the number of singular values 
used in the factorization to reduce the size of the model while minimizing the loss of information. The number of 
singular values was determined empirically based on a trade-off between the size of the model and its accuracy 
on validation data. We then reconstructed the original weight matrix using the truncated SVD matrices, effectively
 reducing the size of the weight matrix. Finally, we reshaped the weight matrix back into its original tensor shape
  to restore the weight tensor of each convolutional layer in the model.

To evaluate the effectiveness of the SVD optimization technique, we compared the performance of the optimized model 
with that of the original unoptimized model using publicly available dataset\cite{street-work_dataset}. We also
 varied the number of singular values used in the SVD factorization and evaluated the impact on model size and accuracy.
\begin{equation}
\mathbf{W}_{m\times n} \approx \mathbf{U}_{n\times k} \mathbf{\Sigma}_{k\times k}\mathbf{V}^\top_{k\times n}
\end{equation}

where $\mathbf{W}$ is the weight tensor of a convolutional layer in the model, reshaped into a 2D matrix with dimensions
 $m \times n$, $\mathbf{U}$ is an $m \times k$ matrix, $\mathbf{\Sigma}$ is a $k \times k$ diagonal matrix containing
  the singular values, and $\mathbf{V}^\top$ is a $k \times n$ matrix. The number $k$ is chosen to be smaller than 
  both $m$ and $n$, and represents the rank of the truncated SVD. 

As demonstrated in Table \ref{tab:SVD-Params},  the number of parameters in the Original Weight is \textbf{$OIK^{2}$}, 
while in the case of U+S+V, it is \textbf{$R(IK^{2}+1+O)$}. Typically, \textbf{$R(IK^{2}+1+O)$} is smaller than 
\textbf{$OIK^{2}$}, particularly when the rank of SVD decreases.

\begin{table}[]
\centering
\caption{The Number of Parameters Difference between original Model Weight and Low-Rank SVD}
\label{tab:SVD-Params}
\begin{tabular}{ccccccc}
\hline
\multicolumn{1}{|c|}{\textbf{W}}          & \multicolumn{1}{c|}{\textbf{Origin}}                 & \multicolumn{1}{c||}{\textbf{Reshaped}}                       & \multicolumn{1}{c|}{\textbf{U}}                              & \multicolumn{1}{c|}{\textbf{S}}       & \multicolumn{1}{c|}{\textbf{V}}          & \multicolumn{1}{c|}{\textbf{U+S+V}}                                  \\ \hline
\multicolumn{1}{|c|}{\textbf{Shape}}      & \multicolumn{1}{c|}{\textbf{${[}O, I, K, K{]}$}}       & \multicolumn{1}{c||}{\textbf{${[}IK^{2}, O{]}$} } & \multicolumn{1}{c|}{\textbf{${[}IK^{2}, R{]}$}} & \multicolumn{1}{c|}{\textbf{${[}R{]}$}} & \multicolumn{1}{c|}{\textbf{${[}O, R{]}$}} & \multicolumn{1}{c|}{\textbf{${[}IK^{2}+1+O, R{]}$}} \\ \hline
\multicolumn{1}{|c|}{\textbf{Parameters}} & \multicolumn{1}{r|}{\textbf{$OIK^{2}$}} & \multicolumn{1}{r||}{\textbf{$OIK^{2}$}}         & \multicolumn{1}{r|}{\textbf{$RIK^{2}$}}         & \multicolumn{1}{r|}{\textbf{$R$}}       & \multicolumn{1}{r|}{\textbf{$OR$}}         & \multicolumn{1}{r|}{\textbf{$R(IK^{2}+1+O)$}}           \\ \hline
\multicolumn{1}{l}{}                      &                                                      &                                                              &                                                              &                                       &                                          &                                                                      \\
\multicolumn{7}{l}{I : Input Channels, O : Output Channels, K : Kernel Size, R : Rank of SVD}                                                                                                                                                                                                                                                                                                                                           
\end{tabular}
\end{table}

\noindent{\bf Hyperparameters}\\
Hyperparameters are parameters that are not learned during the training of a machine learning model, but rather
 set by the practitioner prior to training. These parameters govern the behavior of the learning algorithm and 
 can significantly affect the performance of the model. In YOLOv7, there are several hyperparameters that can be 
 tuned to achieve optimal performance. Among these hyperparameters, some of the most significant ones include 
 the batch size, learning rate, and weight decay.

Batch size is the number of training examples used in each iteration of the optimization algorithm during training 
of a machine learning model. It determines how many samples from the training dataset are processed at once before 
the model weights are updated. The choice of batch size can affect the training process and the resulting model 
performance. A larger batch size can speed up training but may require more memory and may lead to poorer generalization 
performance. A smaller batch size may take longer to train but may lead to better generalization performance and smoother convergence.

Learning rate is another important hyperparameter that controls the step size at which the optimization algorithm updates 
the weights of the model during training. It determines the magnitude of the updates to the model weights in each iteration. 
A high learning rate can cause the optimization algorithm to overshoot the optimal weights, leading to unstable training
 and poor performance. A low learning rate can cause the optimization algorithm to converge too slowly, leading to longer 
 training times and potentially suboptimal performance.

Weight decay is a regularization technique that is used to prevent overfitting in machine learning models. It involves 
adding a penalty term to the loss function that encourages the model to learn simpler and more generalizable patterns 
in the data. The strength of the penalty term is controlled by the weight decay hyperparameter. A high weight decay 
can lead to overly simplified models that underfit the training data, while a low weight decay can lead to overfitting 
and poor generalization performance.

The appropriate choice of learning rate and weight decay depends on the specific dataset and network architecture being 
used, as well as the optimization algorithm and other hyperparameters. In practice, it is often necessary to perform a 
grid search or other hyperparameter tuning techniques to identify the optimal values for these hyperparameters. By 
carefully selecting these values, a neural network can achieve better performance and generalization capabilities.

During our experimentation, we evaluated several transfer learning approaches by varying the batch size during training.
 We tested batch sizes of 1, 16, 32, and 64, and found that the model achieved the highest mean average precision (mAP) 
 when the batch size was set to 32. Based on these results, we selected a batch size of 32 for our study because it
  achieved the best mAP and exhibited appropriate training speed and generalization performance. However, since our 
  research topic did not involve hyperparameter tuning, we used the default values of 'hyp.scratch.tiny.yaml' 
  provided by YOLOv7's GitHub repository, which includes the default values of lr0:0.01, lrf:0.01, weight decay:0.0005, among other hyperparameters.

\section{Results}
\begin{figure}[t!]
    \centering
    \includegraphics[width=0.48\textwidth]{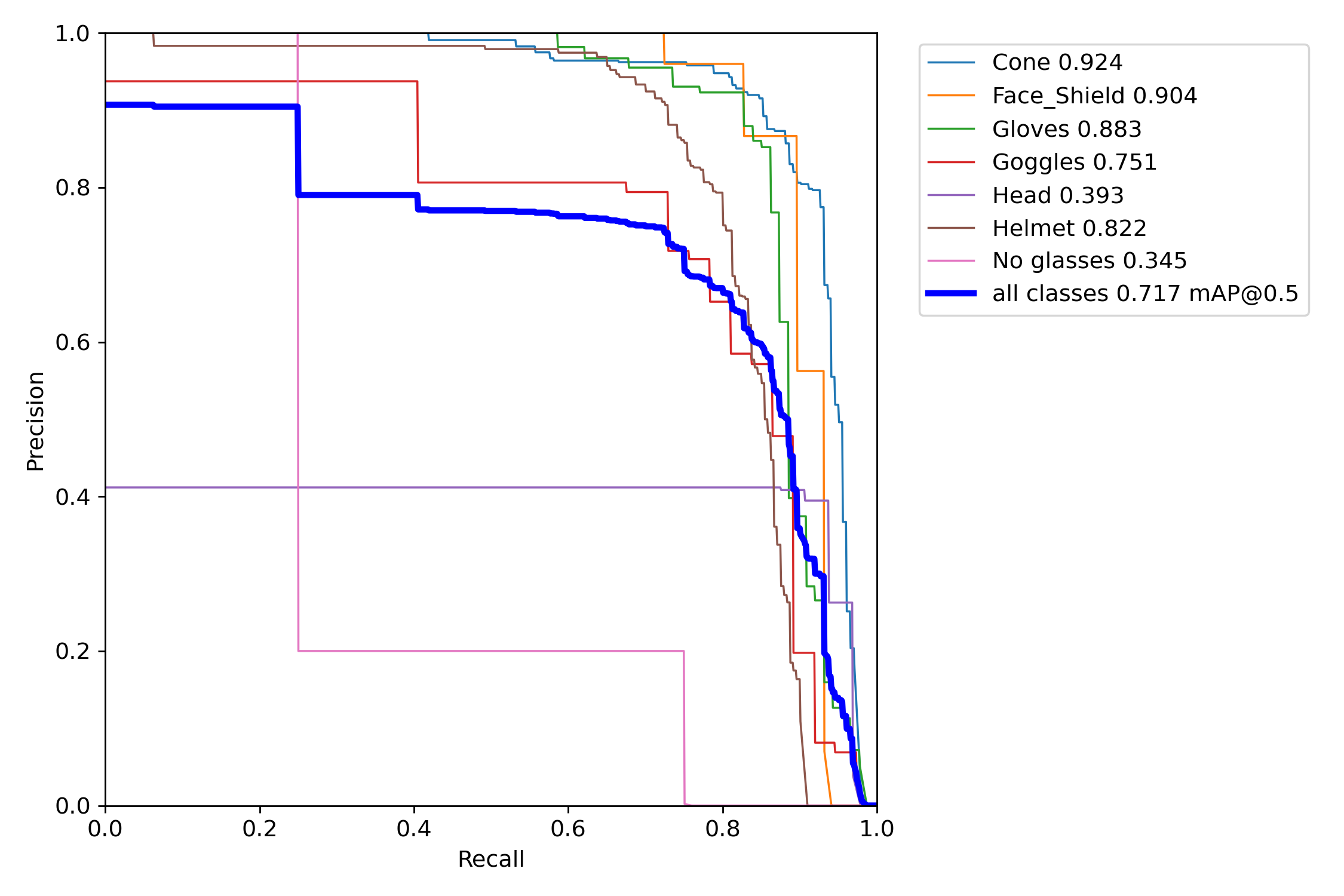}
    \caption{PR curve of unoptimized model on test dataset.}
    \label{fig:original_pr_curve}
\end{figure}

\begin{table}[t!]
\centering
\caption{Frame Rates and mAP50 of Models with Weight Pruning and SVD}
\label{tab:frame_rates}
\begin{tabular}{c|c|c|c}
\hline
Model & Frame Rate (FPS) & mAP@50 & Weight Size (MB)\\ \hline
Original & 1.50 & 0.717 & 12.3\\
SVD only & 1.44 & 0.717 & 12.1\\
 & - & 0.677 & 9.92\\
Weight pruning + SVD & 1.48 & 0.724 & 12.1\\ \hline
\end{tabular}
\end{table}

\begin{figure}[b]
    \centering
    \subfigure[SVD Optimized Model]{
        \includegraphics[width=0.48\textwidth]{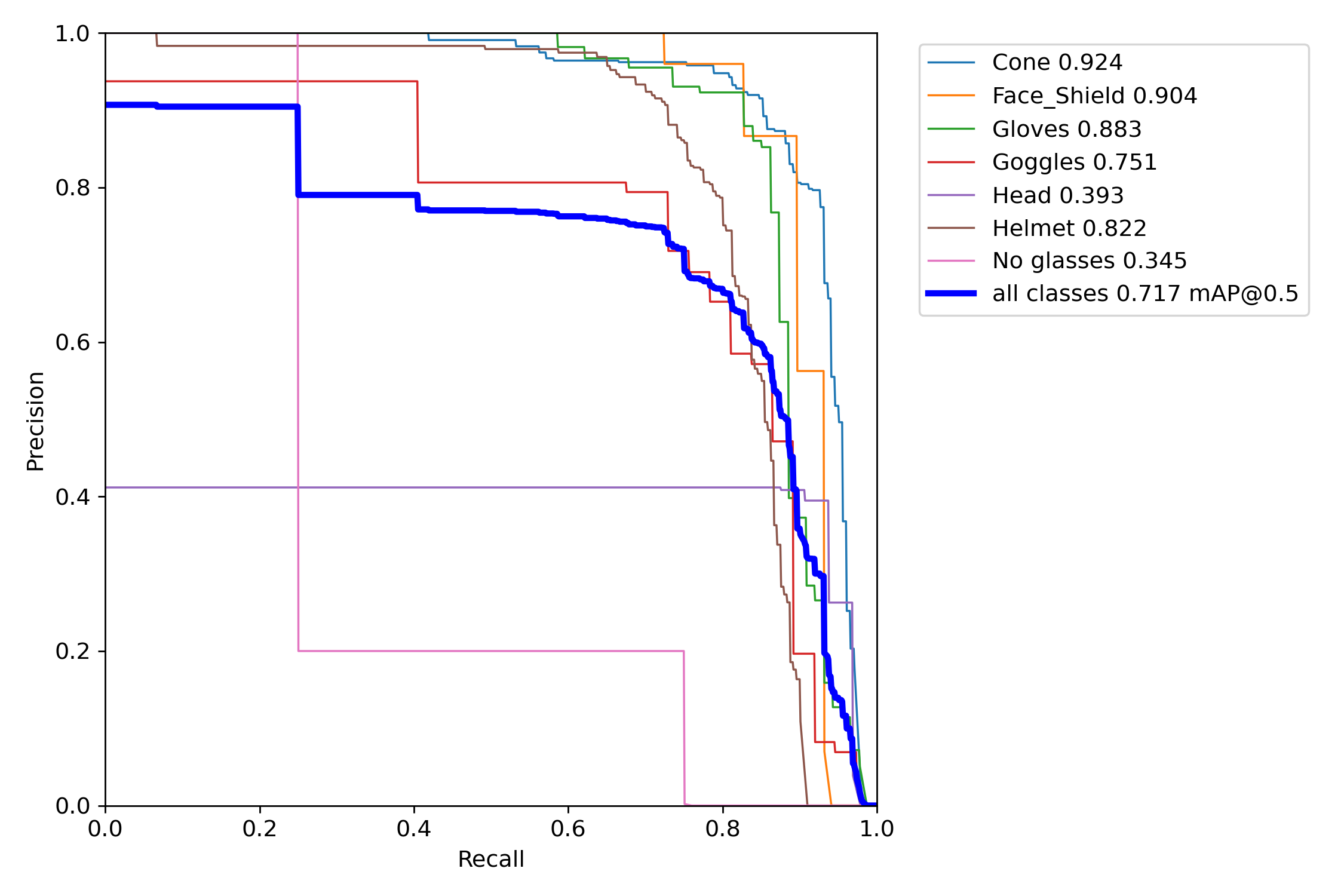}
        \label{fig:svd_pr_curve}
    }
    \subfigure[SVD + Weight Pruned Model]{
        \includegraphics[width=0.48\textwidth]{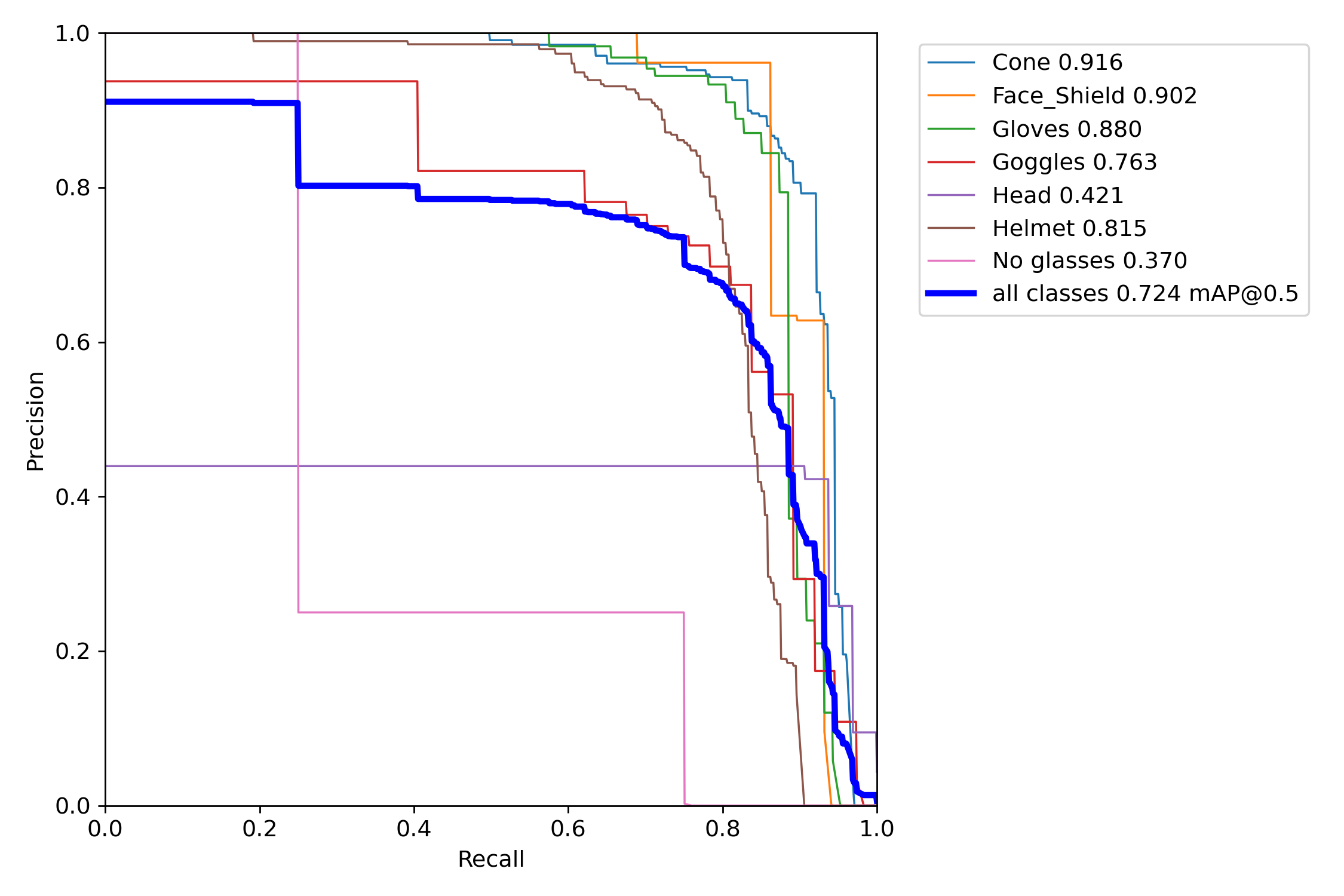}
        \label{fig:svd_pruned_pr_curve}
    }
    \caption{Precision-Recall curves for comparison between SVD optimized model and SVD + weight pruned model.}
    \label{fig:pr_curves}
\end{figure}

Our experiments on standard object detection benchmarks demonstrate the effectiveness of weight pruning and singular value decomposition (SVD) for optimizing the performance of deep learning models. Table~\ref{tab:frame_rates} summarizes the quantitative evaluation of the pruned and SVD-optimized models, as well as the comparison of their performance with the original unoptimized model. We observe that the SVD-optimized model achieves similar mAP@50 compared to the original model, but with a slightly lower frame rate. On the other hand, the weight pruning + SVD model outperforms both the original and SVD-optimized models in terms of mAP@50, while maintaining a comparable frame rate.

Furthermore, we analyze the trade-offs between model size, speed, and accuracy. We note that the weight pruning + SVD model has a similar weight size as the SVD-optimized model, which is significantly smaller than the original model. This reduction in model size is important for applications where storage and memory are limited. In terms of speed, the weight pruning + SVD model has a slightly lower frame rate than the original model, but still maintains a real-time detection capability. These results demonstrate the potential of using pruning and SVD techniques to optimize deep learning models for resource-constrained applications.

The precision-recall curves in Figure~\ref{fig:pr_curves} provide a visual representation of the performance of the object detection models. The PR curve for the original model in Figure~\ref{fig:original_pr_curve} shows that it achieves high precision at low recall values, but its performance drops as the recall increases. In contrast, the PR curves for the SVD-optimized and weight pruning + SVD models in Figure~\ref{fig:pr_curves}(a) and (b), respectively, show that these models achieve better balance between precision and recall, resulting in higher mAP@50 values. Overall, our results highlight the potential of using pruning and SVD techniques to optimize deep learning models for real-world applications.

\section{Discussion}

The results of our study demonstrate that by using weight 
pruning and SVD optimization techniques, we were able to 
reduce the memory footprint of the model while balancing the speed and accuracy of the model. 
The optimized models achieved comparable or even better mAP50 scores 
on the Roboflow dataset compared to the original unoptimized model,
while running with smaller weight sizes. These results could have 
important implications for real-world deployment on resource-constrained 
devices like the Raspberry Pi, where computational resources and memory 
are limited.

Our study is not the first to explore techniques for optimizing 
object detection models for resource-constrained devices. 
Other approaches include quantization, distillation, and network 
compression, among others \cite{polino2018model}. While each of 
these techniques has its own advantages and limitations, 
our results demonstrate that weight pruning and SVD optimization 
can be effective methods for reducing the size and computational
 complexity of object detection models while maintaining or 
 improving their accuracy.

While our study shows promising results, there is still room for improvement and further research in this area. One possible direction for future research is to explore the combination of different optimization techniques to achieve even better results. For example, weight pruning and quantization could be combined to further reduce the size and computational complexity of models. Another area for future research is to investigate the transferability of optimized models across different datasets and object detection tasks.

Additionally, it would be interesting to explore the impact of these optimization techniques on other types of computer vision models, such as image classification or semantic segmentation. Finally, the development of new optimization techniques specifically designed for resource-constrained devices could lead to further improvements in the performance and efficiency of computer vision models in real-world applications.

While we observed no significant increase in FPS with the weight pruning method, if future experiments are conducted with the Raspberry Pi in headless mode (no GUI), there could be a meaningful increase in FPS due to the decrease in bottlenecks.

\section{Conclusion}
Optimization techniques such as weight pruning and singular value decomposition (SVD) can have a significant impact on the real-world applications of object detection models. These real-time inference models are fundamental to computer vision tasks such as autonomous driving, surveillance systems, and medical imaging. The size and computational complexity of these models are critical factors that affect the speed and efficiency of the entire system.

In this study, we evaluated the effectiveness of weight pruning and SVD on the popular YOLO object detection model. We tested the optimized models on a Raspberry Pi platform, which is a popular low-power embedded system used in many real-world applications.

Our results showed that using weight pruning and SVD techniques, we were able to achieve similar performance with smaller model sizes. However, the pace of inference did not improve as expected, possibly due to the limited computing resources of the Raspberry Pi platform. By experimenting with various batch sizes during training, we found that a batch size of 32 achieved the highest average precision.

Further research can focus on quantization and fine-tuning hyperparameters to further optimize these models for real-world applications. The application of these techniques to other popular object detection models could also be explored, as well as the use of more powerful hardware for inference.

Overall, the use of optimization techniques such as weight pruning and SVD 
is a promising approach to reduce the size and computational complexity 
of object detection models, enabling their deployment in 
resource-constrained environments, and improving their efficiency.

\section*{Acknowledgments}
The work presented in this report was conducted as part of the Yonsei-Roboin project for the 2nd semester, 2022. We would like to express our gratitude to Roboin-Yonsei for providing us with the opportunity to conduct this research. We would like to extend our thanks to our project team members for their valuable contributions and collaboration. We would also like to thank the OpenAI team for providing access to the ChatGPT language model, which was used in this research. Finally, we would like to acknowledge the support and encouragement of our friends and families throughout this project.

\bibliographystyle{unsrt}  
\bibliography{main}

\end{document}